\documentclass[sn-mathphys,Numbered]{sn-jnl}

\usepackage{graphicx}%
\usepackage{multirow}%
\usepackage{amsmath,amssymb,amsfonts}%
\usepackage{amsthm}%
\usepackage{mathrsfs}%
\usepackage[title]{appendix}%
\usepackage{xcolor}%
\usepackage{textcomp}%
\usepackage{manyfoot}%
\usepackage{booktabs}%
\usepackage{algorithm}%
\usepackage{algorithmicx}%
\usepackage{algpseudocode}%
\usepackage{listings}%
\usepackage{float}

\theoremstyle{thmstyleone}%
\theoremstyle{thmstyletwo}%

\theoremstyle{thmstylethree}%

\begin{document}

\title[High-Performance Fine Defect Detection in Artificial Leather Using Dual Feature Pool Object Detection]{High-Performance Fine Defect Detection in Artificial Leather Using Dual Feature Pool Object Detection}


\author[1,2,4,5]{\fnm{Lin} \sur{Huang}}\email{h72001346@163.com}

\author*[1]{\fnm{Weisheng} \sur{Li}}\email{liws@cqupt.edu.cn}

\author[2]{\fnm{Yujuan} \sur{Tan}}\email{tanyujuan@cqu.edu.cn}

\author[3]{\fnm{Linlin} \sur{Shen}}\email{llshen@szu.edu.cn}

\author[2]{\fnm{Jing} \sur{Yu}}\email{20201401010@cqu.edu.cn}

\affil*[1]{\orgname{Chongqing University of Posts and Telecommunications}, \orgaddress{\street{No.2, Chongwen Road, Nan'an District}, \city{Chongqing}, \postcode{400065}, \country{China}}}

\affil[2]{\orgname{Chongqing University}, \orgaddress{\street{No. 55, University Town South Road, Gaoxin District}, \city{Chongqing}, \postcode{401331}, \country{China}}}

\affil[3]{\orgname{Shenzhen University}, \orgaddress{\street{No.3688 Nanshan Avenue, Nanshan District}, \city{Shenzhen}, \postcode{518061},  \country{China}}}

\affil[4]{\orgname{Inspur Yunzhou Industrial Internet Co., Ltd}, \orgaddress{\street{No.1036 Langchao Road, Lixia District}, \city{Jinan}, \postcode{250101}, \state{Shandong}, \country{China}}}

\affil[5]{\orgname{Guoqi Zhimo (Chongqing) Technology Co., Ltd.}, \orgaddress{\street{5th Floor, Building B15, Xiantao Data Valley, Yubei District}, \city{Chongqing}, \postcode{401122},\country{China}}}


\abstract{In this study, the structural problems of the YOLOv5 model were analyzed emphatically. Based on the characteristics of fine defects in artificial leather, four innovative structures, namely DFP, IFF, AMP, and EOS, were designed. These advancements led to the proposal of a high-performance artificial leather fine defect detection model named YOLOD. YOLOD demonstrated outstanding performance on the artificial leather defect dataset, achieving an impressive increase of 11.7\% - 13.5\% in AP$_{50}$ compared to YOLOv5, along with a significant reduction of 5.2\% - 7.2\% in Average Error Detection (AE). 
Furthermore, YOLOD demonstrated outstanding performance on the comprehensive MS-COCO dataset, showcasing an improvement ranging from 0.4\% to 2.6\% in AP when compared to YOLOv5. Notably, it achieved a significant boost of 2.5\% and 4.1\% in AP$_S$ compared to YOLOX-L and YOLOv4-CSP, respectively. These results demonstrate the superiority of YOLOD in both artificial leather defect detection and general object detection tasks, making it a highly efficient and effective model for real-world applications.}




\maketitle

\section{Introduction} 

\begin{figure}[H]
\centering
\includegraphics[height=3.5cm]{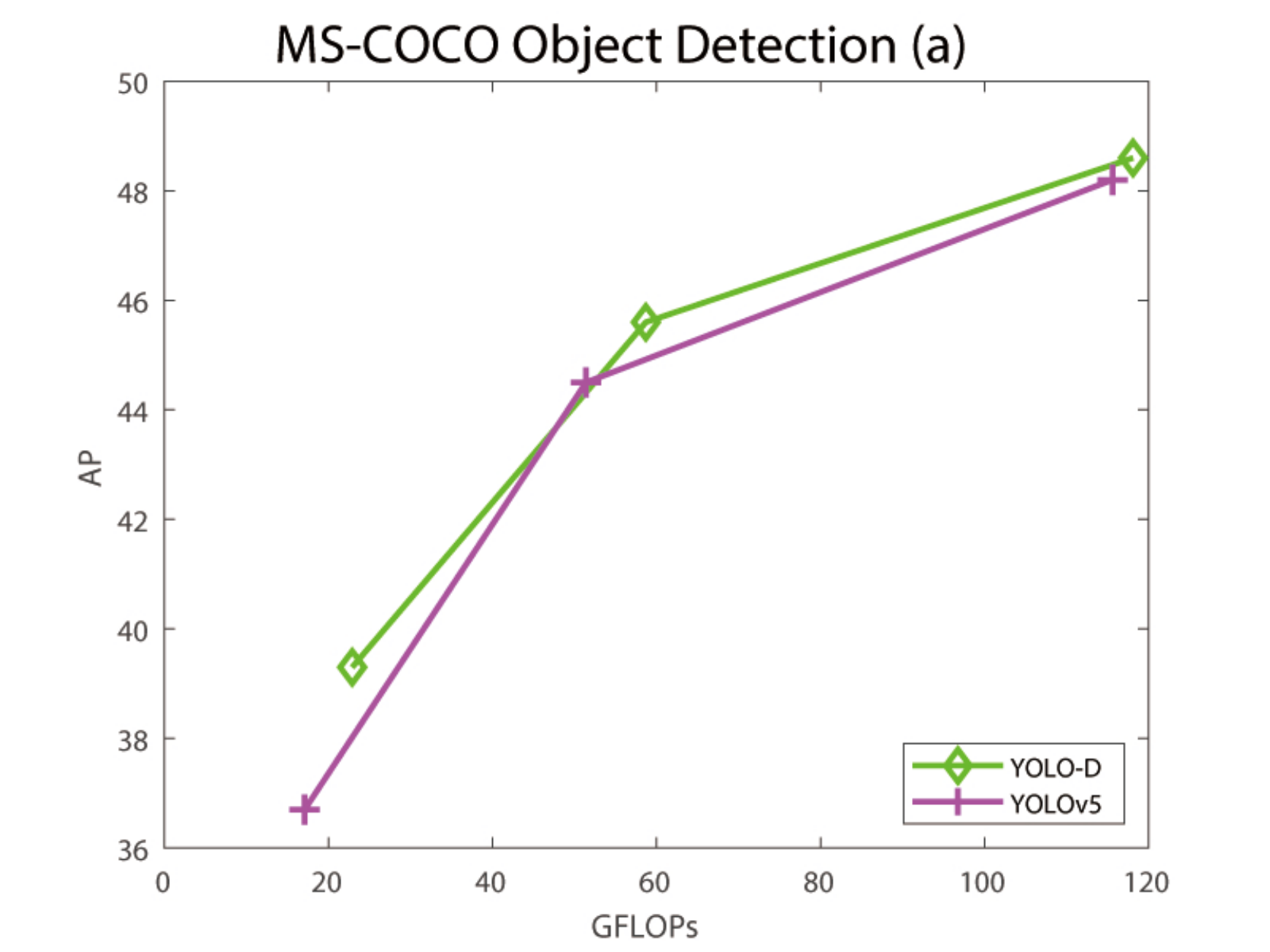}
\includegraphics[height=3.5cm]{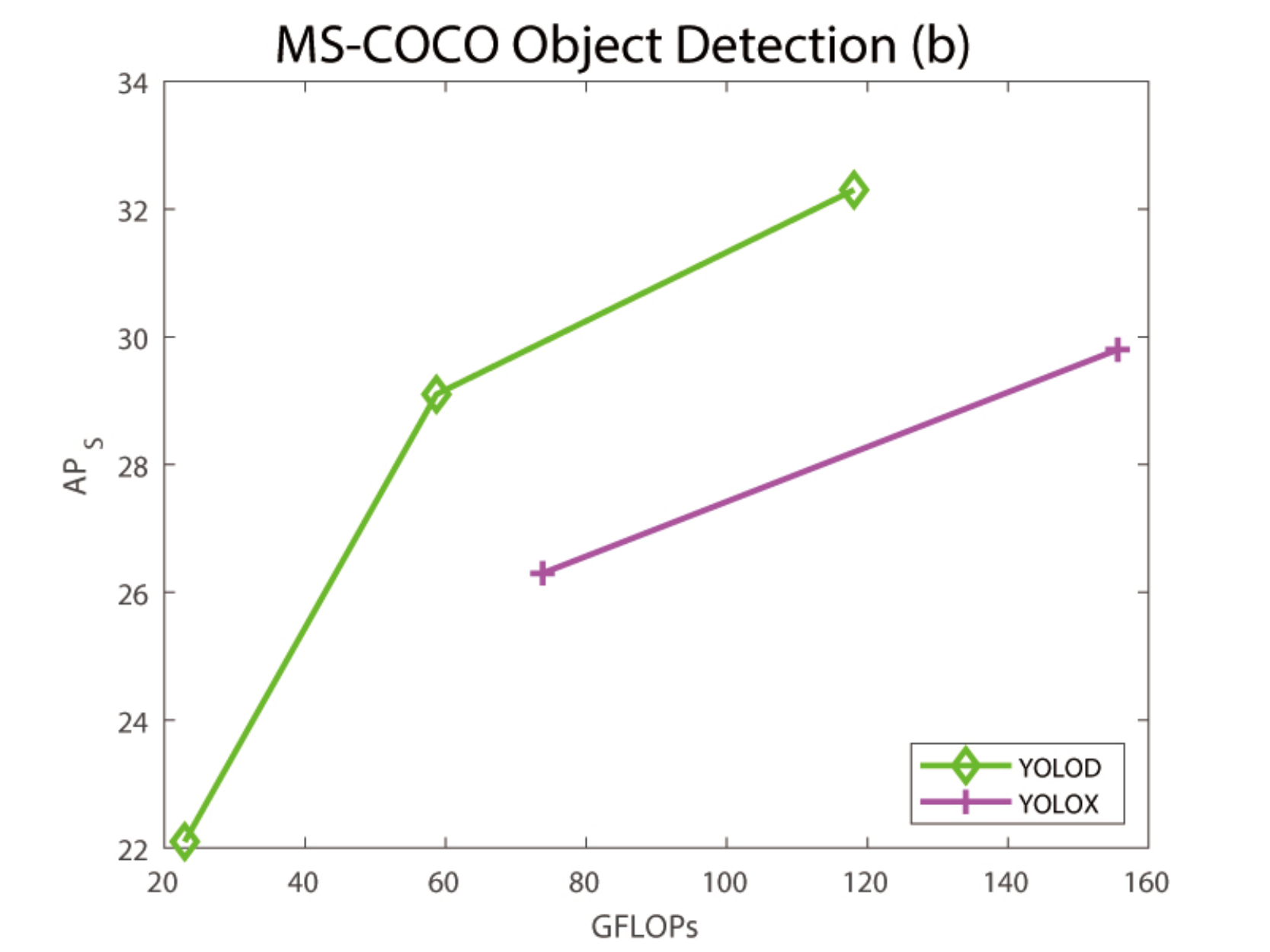}
\includegraphics[height=3.5cm]{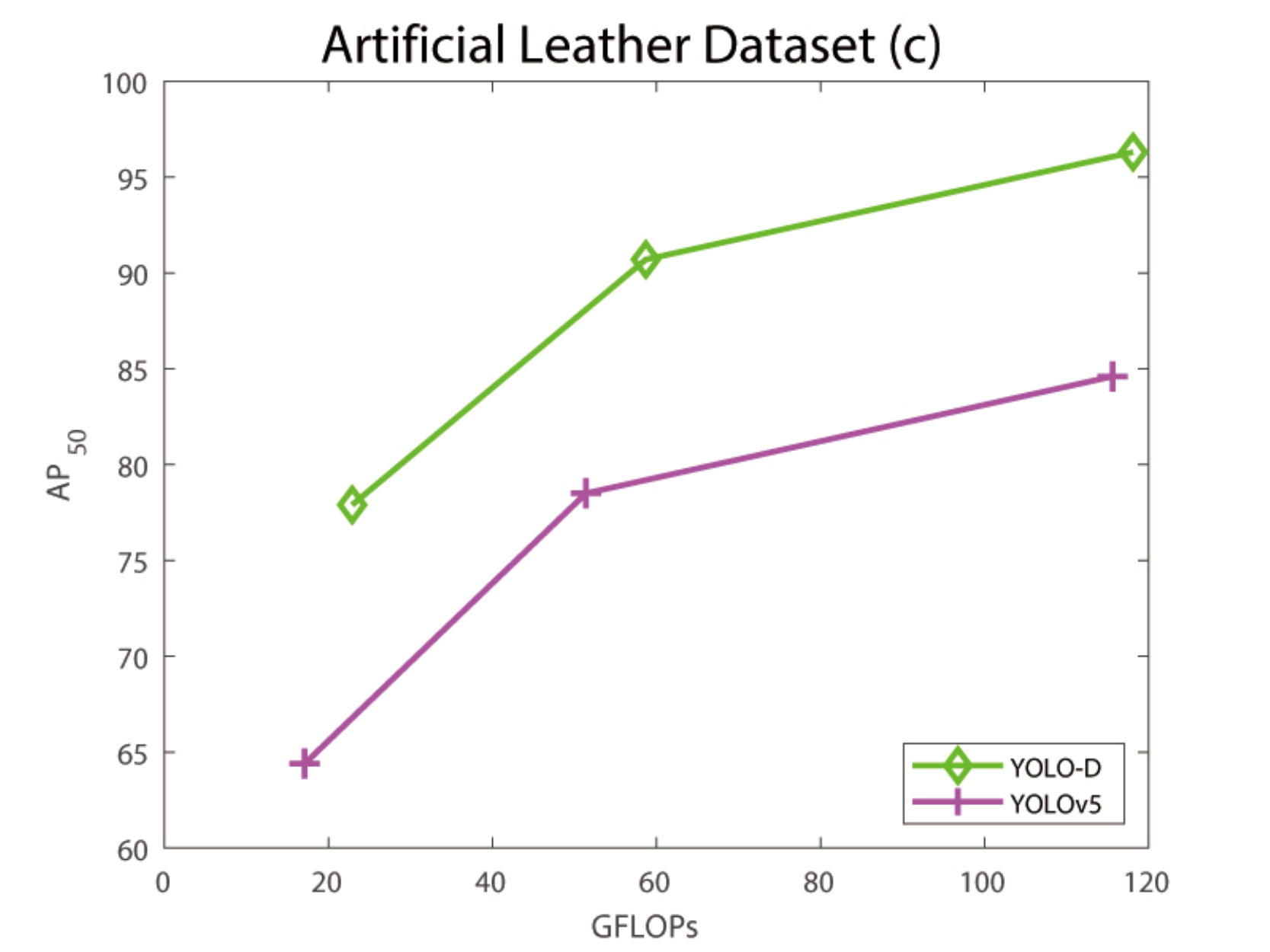}
\includegraphics[height=3.5cm]{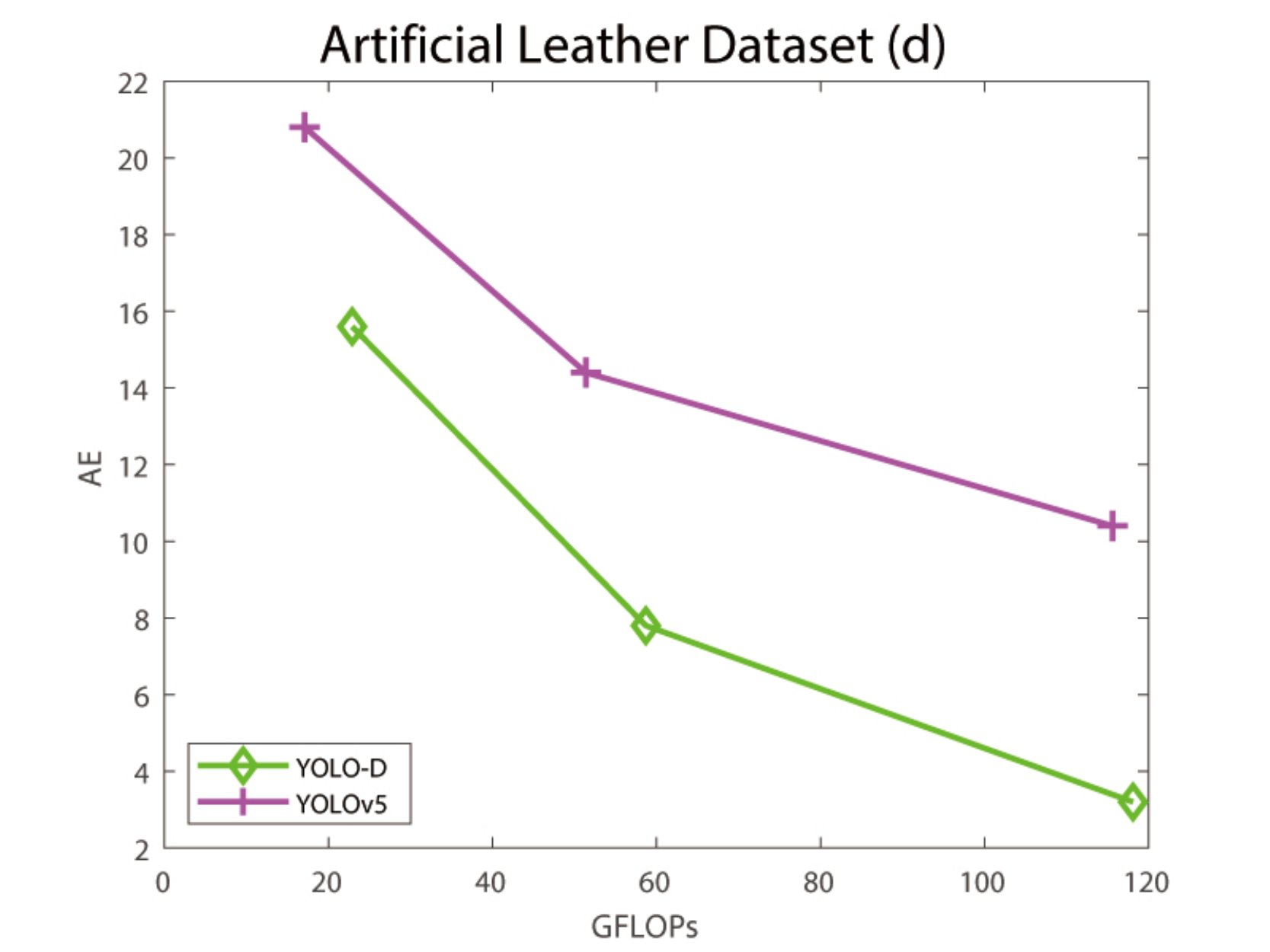}
\caption{(a) illustrates the comparison of AP on MS-COCO between the proposed YOLOD and YOLOv5. (b) illustrates the comparison of AP$_S$ on MS-COCO between the proposed YOLOD and YOLOX. (c) illustrates the comparison of AP$_{50}$ on ALD between the proposed YOLOD and YOLOv5. (d) illustrates the comparison of AE on ALD between the proposed YOLOD and YOLOv5.}
\label{fig:t}
\end{figure}

Artificial leather, also known as synthetic leather, has gained widespread popularity in various industries due to its cost-effectiveness and eco-friendly attributes. However, ensuring its quality and identifying fine defects remain significant challenges. Traditional defect detection of artificial leather primarily relies on human inspection, which poses several limitations in terms of detection efficiency and accuracy. The shortcomings of human visual inspection are twofold. Firstly, it cannot meet the demands of long-term and high-speed production lines, as human vision is prone to fatigue and distraction. As inspection time increases, not only does the efficiency and accuracy of workers decrease, but prolonged exposure to bright lighting conditions can also negatively impact their health. Secondly, human evaluation of defects involves a degree of subjectivity, making it difficult to maintain consistent standards. This inconsistency largely stems from variations in the workers' individual learning abilities and the extent of their accumulated experience. These individual differences can lead to instability in product quality assessments, causing direct economic losses. Therefore, there is an urgent need for alternative technologies to replace manual defect detection and ensure stable product quality.


﻿

With the rise of deep learning, numerous advanced machine vision algorithms based on deep learning have emerged, effectively addressing the challenges posed by the irregular and random patterns of artificial leather and improving defect detection rates. Object detection algorithms have been primarily categorized into two major research directions: one-stage object detection and two-stage object detection. While two-stage object detection\cite{maskrcnn,cascadercnn,rfh,fastrcnn,fasterrcnn,rfcn,lrcnn} outperforms one-stage object detection\cite{yolo1,yolo2,yolo3,yolo4,syolo4,yolo5,yolopp,yolopp2,yolof,yolox,yolo7,focal,efficientdet,asff,ssd} in terms of AP, the latter excels in inference speed and is thus highly suitable for real-time applications in various industries. Notably, the YOLO(You Only Look Once) model, along with SSD\cite{ssd}, stands as one of the most representative detectors among the algorithms of one-stage object detection. The YOLO series of algorithms has evolved from version 1 to version 8\cite{yolo1,yolo2,yolo3,yolo4,syolo4,yolo5,yolopp,yolopp2,yolof,yolox,yolo7}, and inspires a number of variants such as pp-YOLOv1\cite{yolopp}, pp-YOLOv2\cite{yolopp2}, YOLOF\cite{yolof}, and YOLOX\cite{yolox}. These algorithms introduced a wealth of innovative ideas and development paths for advancing one-stage object detection. Among these algorithms, real-time object detection technologies\cite{yolo1,yolo2,yolo3,yolo4,syolo4,yolo5,yolopp,yolopp2,yolof,yolox,yolo7,focal,efficientdet,asff,ssd,fabric,baimobile}  have seen rapid development, focusing not only on improving detection accuracy but also on enhancing detection efficiency. These advancements open up new possibilities for applying deep learning to the detection of defects in artificial leather\cite{al1,al2,al3,al4,al5}.





\begin{figure}[H]
\centering
\includegraphics[height=6.5cm]{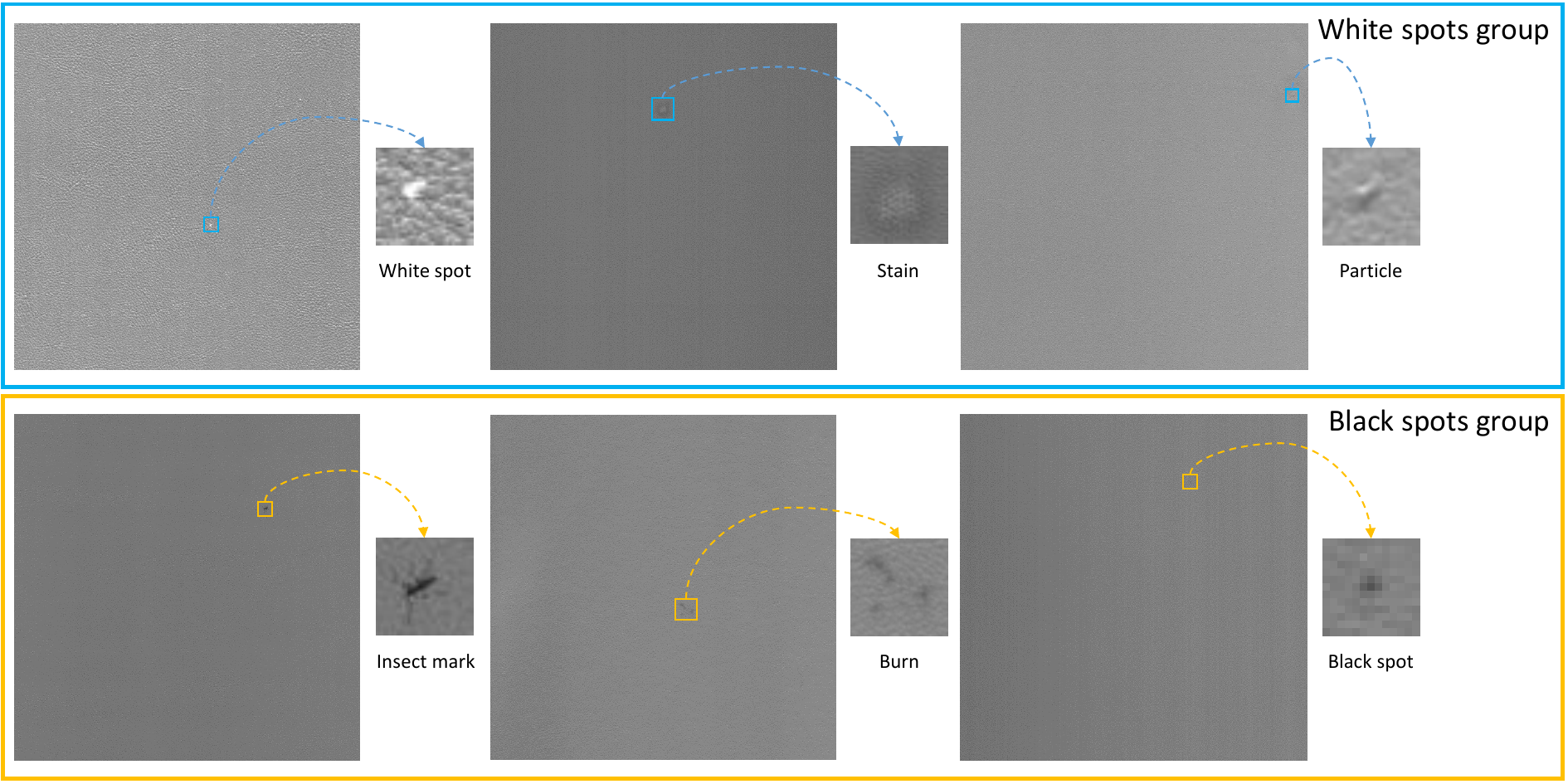}
\caption{Two types of artificial leather defects are black spots and white spots.}
\label{fig:ald}
\end{figure}

The defects in synthetic leather can be primarily categorized into two major types: linear defects, including scratches, creases, fine stripes, etc., and point defects, such as white spots, black spots, stains, insect marks, oil stains, impurities, particles, burns, etc. Since detection of large-sized defects is relatively straightforward, this paper focuses solely on the more challenging task of detecting small-sized defects with deep learning algorithms. Specifically, the defects of interest include black spots, white spots, insect marks, impurities, particles, burns, and stains. These defects, when imaged using illumination and industrial grayscale cameras, can be broadly categorized into two groups: black spots and white spots (Fig.\ref{fig:ald}). The baseline model adopted in this paper is YOLOv5\cite{yolo5}. YOLOv5 is widely applied in industrial fields due to its efficient inference speed and relatively high detection rate (AP).

This paper begins by evaluating the model on the public dataset MS-COCO\cite{mscoco}, where YOLOD demonstrates an AP\cite{mscoco} improvement ranging from \textbf{0.4\%} to \textbf{2.6\%} over YOLOv5(Fig.\ref{fig:t}(a)). Additionally, as YOLOv5 does not report small object average precision (AP$_S$)\cite{mscoco}, this study performs a comparative analysis of YOLOD's small object detection AP$_S$ with that of YOLOX-L\cite{yolox} and YOLOv4-CSP\cite{yolo4}, which shows a noticeable increases from \textbf{2.5\%} to \textbf{4.1\%}(Fig.\ref{fig:t}(b)), respectively. Moreover, this paper evaluates the model on the Artificial leather dataset collected from a real industry project. In this evaluation, YOLOD exhibits an impressive \textbf{11.7\% - 13.5\%} increase in AP$_{50}$ compared to YOLOv5(Fig.\ref{fig:t}(c)), along with a significant \textbf{5.2\% - 7.2\%} reduction in AE(Fig.\ref{fig:t}(d)). In summary, the contributions of this paper can be outlined as follows:

1. This paper proposes a novel neck network called the "dual feature pool" to effectively tackle the challenge of small object detection caused by imbalanced depth within the neck network, built upon the FPN+PAN structure. The dual feature pool structure not only addresses the issue and enhances the AP$_S$, but also leads to an overall improvement of the AP. Additionally, the proposed model demonstrates enhanced detection capabilities of small defects in artificial leather.

2. This paper proposes a novel structure called the "Interference Feature Filtering." This structure assesses the quality of channel feature maps at critical nodes and effectively filters out 0.5\% to 5\% of the low-score channel feature maps. This approach effectively reduces the interference of irregular patterns in artificial leather on small defects during the feature decomposition process.

3. This paper has enhanced the technique for incorporating positive samples in the loss function. This improvement facilitates the automatic expansion of the number of positive samples based on the bounding box size of the ground truth.

4. This paper thoroughly analyzed the method of eliminating grid sensitivity (Yolo4\cite{yolo4}). Subsequently, we propose an enhanced approach that involves modifying the slope of the sigmoid curve to achieve better convergence when increasing the scale parameter.

\section{Dataset}
\subsection{Microsoft coco: Common objects in context (MS-COCO)}


Since YOLOv4, one-stage object detection has been evaluated using MS-COCO dataset. YOLOD is also evaluated using the MS-COCO dataset. The training set of MS-COCO train2017 dataset consists of 118,287 images along with corresponding object annotation. The validating set of MS-COCO val2017 dataset contains 5,000 images and corresponding object annotation data, which are mainly utilized for model evaluation. Furthermore, the MS-COCO test2017 dataset includes 40,670 images without corresponding object annotation data, for testing. To evaluate the performance on this dataset, the inference results (JSON\cite{json} format data) are required to be uploaded to the official MS-COCO website\footnote{https://cocodataset.org/} for evaluation, which are assessed using a number of matrics presented in Tab.\ref{table:mscoco}.

\begin{table}[h]
\caption{The performance metrics for evaluating object detection on the MSCOCO dataset}
\label{table:mscoco}
\begin{tabular}{l l}
\hline
{Metrics}&{Description}\\
\hline
Average Precision (AP):& \\
\setlength{\parindent}{2em}$AP$ & AP at IoU=.50:.05:.95 (primary challenge metric) \\
\setlength{\parindent}{2em}$AP^{IoU=.50}$ & AP at IoU=.50 (PASCAL VOC metric) \\
\setlength{\parindent}{2em}$AP^{IoU=.75}$ & AP at IoU=.75 (strict metric)\\
AP Across Scales:& \\
\setlength{\parindent}{2em}$AP^{small}$ & AP for small objects: area \textless $32^2$ \\
\setlength{\parindent}{2em}$AP^{medium}$ & AP for medium objects: $32^2$ \textless area \textless $96^2$ \\
\setlength{\parindent}{2em}$AP^{large}$ & AP for large objects: area \textgreater $96^2$ \\
Average Recall (AR):& \\
\setlength{\parindent}{2em}$AR^{max=1}$ & AR given 1 detection per image \\
\setlength{\parindent}{2em}$AR^{max=10}$ & AR given 10 detections per image \\
\setlength{\parindent}{2em}$AR^{max=100}$ &  AR given 100 detections per image \\
AR Across Scales:& \\
\setlength{\parindent}{2em}$AR^{small}$ & AR for small objects: area \textless $32^2$ \\
\setlength{\parindent}{2em}$AR^{medium}$ & AR for medium objects: $32^2$ \textless area \textless $96^2$ \\
\setlength{\parindent}{2em}$AR^{large}$ & AR for large objects: area \textgreater $96^2$ \\
\hline
\end{tabular}
\end{table}

\subsection{Artificial leather dataset (ALD)}

\begin{figure}[H]
\centering
\includegraphics[height=5cm]{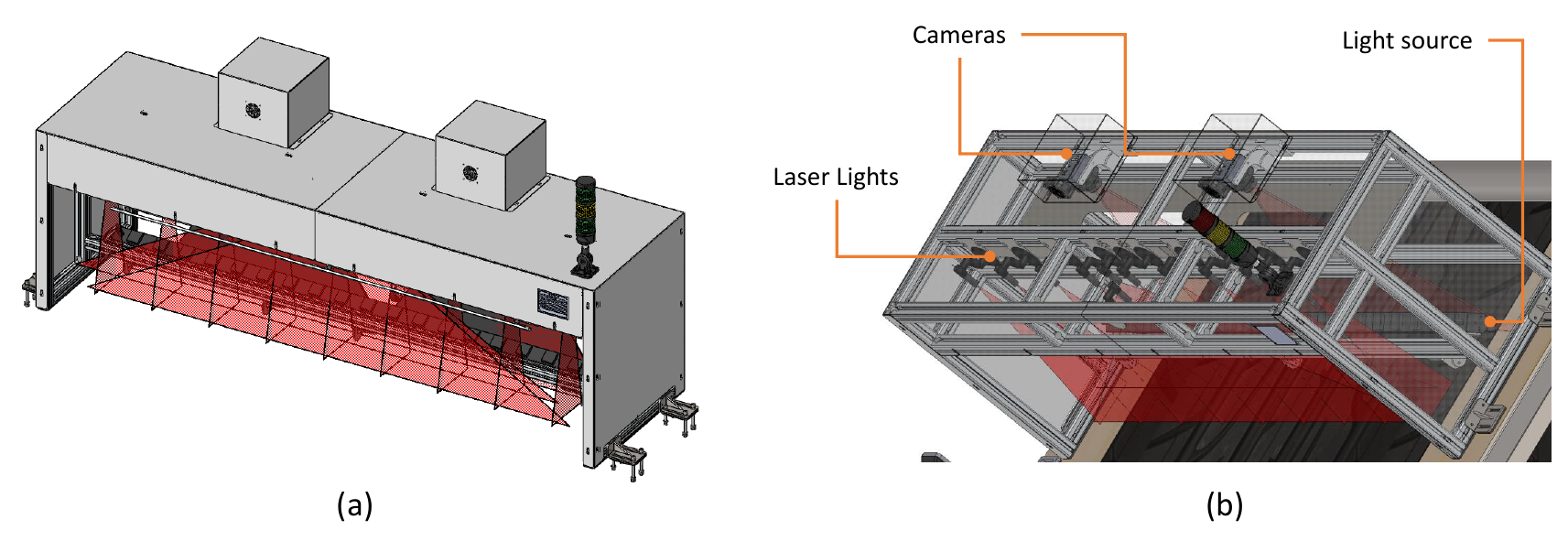}
\caption{Figure (a) presents the external view of the mechanical structure, while Figure (b) shows its perspective view.}
\label{fig:ms}
\end{figure}

\begin{table}[h]
\caption{The performance metrics for evaluating object detection on the ALD dataset}
\label{table:alds}
\begin{tabular}{c c | c c}
\hline
\multicolumn{2}{c|}{\multirow{2}{*}{ Metrics}} & \multicolumn{2}{c}{ Results }\\
 & &Positive& Negative \\
\hline
\multirow{2}{*}{Ground-truth}&True &TP&FN\\
&False &FP&TN\\
\hline
\end{tabular}
\end{table}

The defect dataset of Artificial leather originates from a real project conducted by our company. In the initial data collection phase, we employed a proof-of-concept (POC) device comprising an LED light source, laser lights, cameras, control software, mechanical structure (Fig.\ref{fig:ms}), and computing equipment. The ALD dataset consists of 10,000 images for model training(ALD train2021), and 2,000 images for model evaluation(ALD val2021).  All images were collected using the equipment shown in Fig.2, mounted on the final product inspection machine of an artificial leather manufacturer, and manually screened. The image acquisition setup included one custom light source and two 8K line-scan cameras. The captured images have a resolution of 8192×2048 pixels and were cropped into four 2048×2048 square images, then resized to 1024×1024. The dataset contains only six categories of fine defects, with a total of 12,231 objects, including 2,901 white spots, 1,892 stains, 925 particles, 3,346 black spots, 1,238 insect marks, and 1,929 burn marks.Furthermore, this paper employs standard object detection evaluation metrics for the ALD dataset, as shown in Tab.\ref{table:alds}. In the "results" column, "positive" and "negative" represent predictions of defect and non-defect, respectively. In the "ground truth" column, "True" and "False" indicate whether the actual result in the validation set corresponds to defect or non-defect, respectively. TP (True Positive) denotes cases where both the predicted and actual results are defects. FN (False Negative) represents cases where the prediction is non-defect but the actual result is defect (typically indicating a missed detection). FP (False Positive) refers to cases where the prediction is defect but the actual result is non-defect (typically indicating a false alarm).The calculation process of the standard evaluation metrics is as follows:

\begin{equation}
\begin{aligned}
Precision = \frac{TP}{TP + FP}=\frac{TP}{All Detections}
\end{aligned}
\label{eq:m1}
\end{equation}

\begin{equation}
\begin{aligned}
Recall = \frac{TP}{TP + FN}=\frac{TP}{All Ground Truths}
\end{aligned}
\label{eq:m2}
\end{equation}

Based on the above equations and Tab.\ref{table:alds}, FP can be regarded as detections erroneously labeled as positive, and FN as instances missed during detection. Subsequently, the values of TP, FP, and FN within the evaluation samples are calculated using the IoU\cite{ciou} threshold of 0.5. Precision (p) and recall (r) are then computed, and, following the equations below, the Average Precision at IoU 0.5 (AP$_{50}$\cite{voc}) is calculated:

\begin{equation}
\begin{aligned}
AP_{50} = \sum_{i=0}^{n-1}{(r_{i+1}-r_{i})max(p(r_{i+1}))}
\end{aligned}
\label{eq:ap50}
\end{equation}
Where $n$ represents the number of ground truth instances in the evaluation samples, $r$ denotes recall, $p$ represents precision, and $max$ signifies the maximum precision at the current recall level.

The AP$_{50}$ calculated by Eq.\ref{eq:ap50} primarily addresses a single class. In the case of multiple classes, the calculation needs to be extended to include the following:

\begin{equation}
\begin{aligned}
mAP = \frac{\sum_{i=1}^{K}{AP_i}}{K}
\end{aligned}
\label{eq:map}
\end{equation}
Where $mAP$ denotes the mean AP$_{50}$ across all classes. The mAP is commonly referred to as simply using AP$_{50}$, where the default term AP$_{50}$ typically represents the mean AP$_{50}$ across all classes. AP$_{i}$ represents the AP$_{50}$ value for the specific class, and $K$ signifies the number of classes.

AP$_{50}$ is the most commonly used evaluation metric in object detection. However, in the practical application of artificial leather defect detection, it is crucial to focus not only on the detection performance (AP$_{50}$) but also on the instances of false positives within all detections. This situation can lead to misjudgments regarding the causes of defects in the production process, thereby affecting the smooth operation of the production line. Therefore, we propose a new evaluation metric called Average Error Detection (AE). To compute this metric, it is essential to initially ascertain the count of false positives. The calculation procedure for Error Detection(False Discovery Rate\cite{fdr}) metrics is outlined as follows:

\begin{equation}
\begin{aligned}
Error Detection = \frac{FP}{FP + TP} =\frac{FP}{All Detections} = 1-Precision
\end{aligned}
\label{eq:m3}
\end{equation}

After obtaining the ErrorDetection (e), we summarize the equation for calculating AE as follows:

\begin{equation}
\begin{aligned}
AE = \sum_{i=0}^{n-1}{(r_{i+1}-r_{i})max(e(r_{i+1}))}
\end{aligned}
\label{eq:ae}
\end{equation}
Where $n$ represents the number of ground truth instances in the evaluation samples, $r$ denotes recall, $e$ represents Error Detection, and $max$ signifies the maximum Error Detection at the current recall level.

The AE calculated by Eq.\ref{eq:ae} primarily addresses a single class. In the case of multiple classes, the calculation needs to be extended to include the following:

\begin{equation}
\begin{aligned}
mAE = \frac{\sum_{i=1}^{K}{AE_i}}{K}
\end{aligned}
\label{eq:mae}
\end{equation}
Where $mAE$ denotes the mean AE across all classes. AE$_i$ represents the AE value for the specific class, and $K$ signifies the number of classes. Therefore, following convention, we also refer to mAE as AE.

In this paper, the evaluation of models on the ALD dataset prioritizes the AP$_{50}$ metric. Additionally, considering the significant impact of error detection on product quality in the process of artificial leather defect detection, the AE metric is introduced to assess the models for error detection. Through the evaluation of models using these two metrics, effective identification of models more suitable for artificial leather product defect detection can be achieved.



\section{YOLOD}

\subsection{Dual Feature Pool}

\begin{figure}[H]
\centering
\includegraphics[height=7cm]{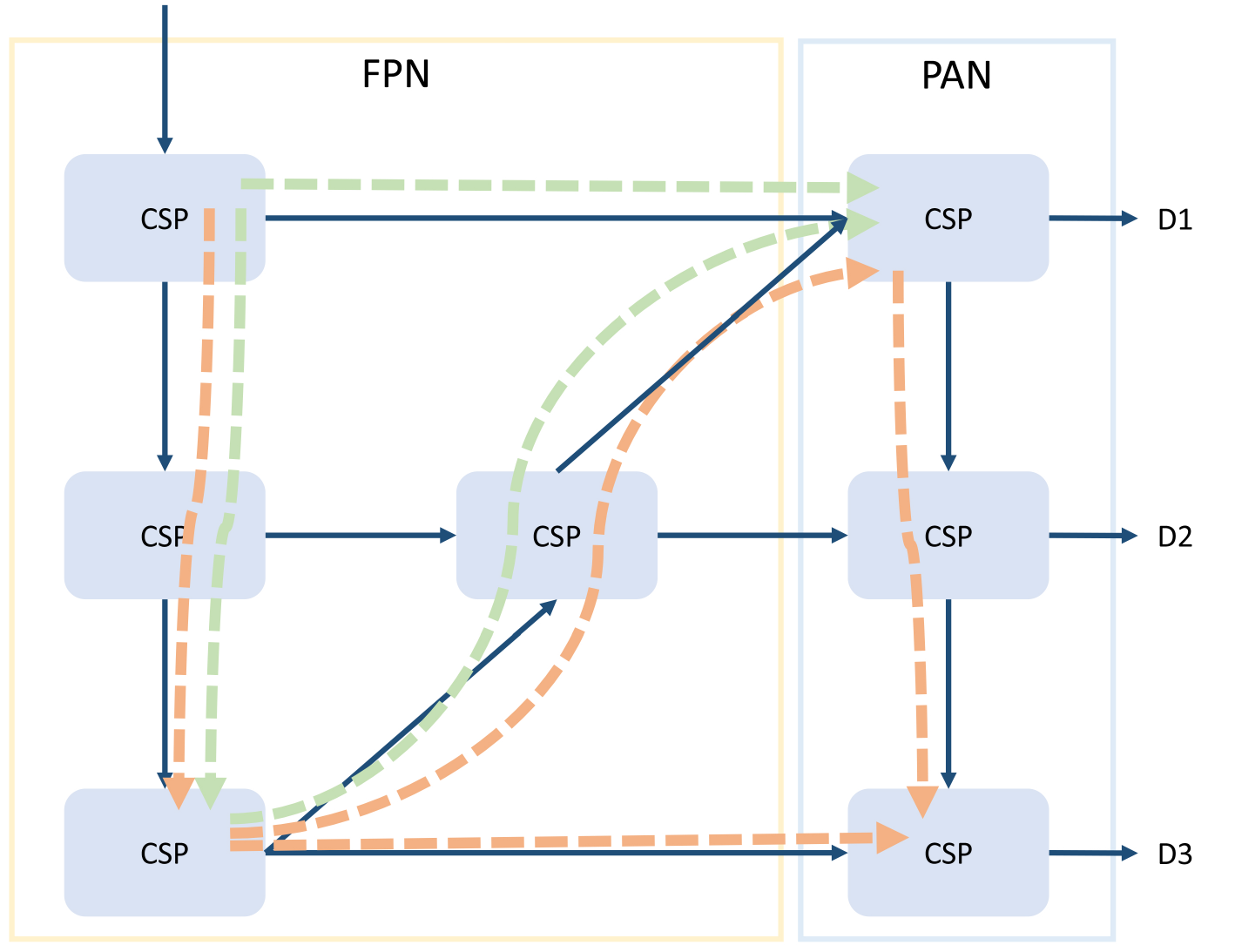}
\caption{The FPN+PAN structure. The orange dashed arrow represents the network path of the detection head D3, while the green dashed arrow represents the network path of the detection head D1. It is evident that the orange network is deeper than the green network. Partial convolution and SPP\cite{spp} are not included from the figure.}
\label{fig:fpn}
\end{figure}

An analysis of the FPN\cite{fpn} + PAN\cite{pan} structure was conducted before proposing the new structure. It was found that in this combined architecture, the detection network's depth for small objects is shallower than that of the network for detecting large objects (Fig.~\ref{fig:fpn}). To improve network detection accuracy, increasing the depth and width of the network becomes essential\cite{efficientnet}. Consequently, the FPN+PAN structure was sub-optimal for the detection of small objects. To address this problem, the objective was to discover a structure that could enhance the depth or width of the small object detection network, thereby improving the accuracy of detecting small objects. Consequently, the dual feature pool structure (DFP) was developed. In this context, the terminology is defined as follows: The features directly output from the backbone network are referred to as "source features," and the features processed by the neck network are termed "processed features."

\begin{figure}
\centering
\includegraphics[height=8cm]{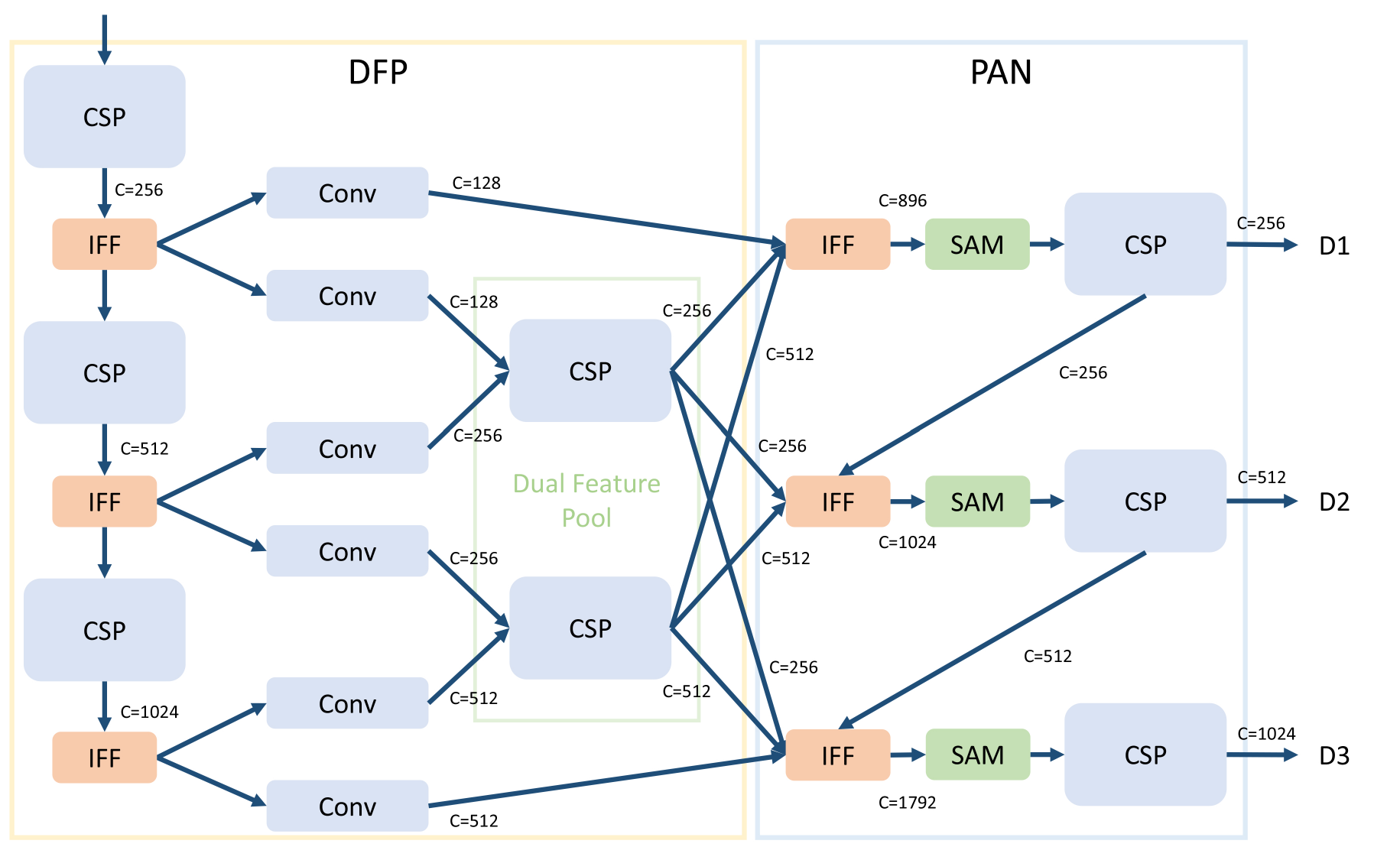}
\caption{The diagram illustrates the DFP+PAN structure. The green box contains two CSP\cite{csp} structures formed by the combination of three-level features. These two CSP structures constitute two feature pools of different scales and output features to the detection head. Partial convolution and SPP are excluded from the figure.}
\label{fig:adfp}
\end{figure}

The network structure is divided into three parts. The first part adopts a structure similar to CSPNet\cite{csp}, which divides three-level source features outputted by the backbone network into two parts through convolution using a half-channel. In the second part, one of the two outputs from each level is fused to form two feature pools (Fig.~\ref{fig:adfp}). The third part fuses another portion of the source features outputted at different levels with the output of the two feature pools. Before inputting each detection head, the fused features are enhanced using Interference Feature Filtering (IFF) and Spatial Attention Module(SAM)\cite{cbam}. By employing this structure, the goal was to expand the network width as much as possible while keeping the network depth unchanged. The DFP structure is comprised of these three parts, which effectively widens the entire neck section, particularly the small object detection network. The two feature pools play a crucial role as they integrate small-medium-scale and medium-large-scale features, offering more feature dimensions to the detection head. Following the fusion in the third part, the dimension of features passed to the detection head increases by 1.75 to 2 times compared to FPN + PAN. The initial intention behind adopting this structure was to expand the width of the small object network.

\subsection{Interference Feature Filtering}

\begin{figure}[H]
\centering
\includegraphics[height=5cm]{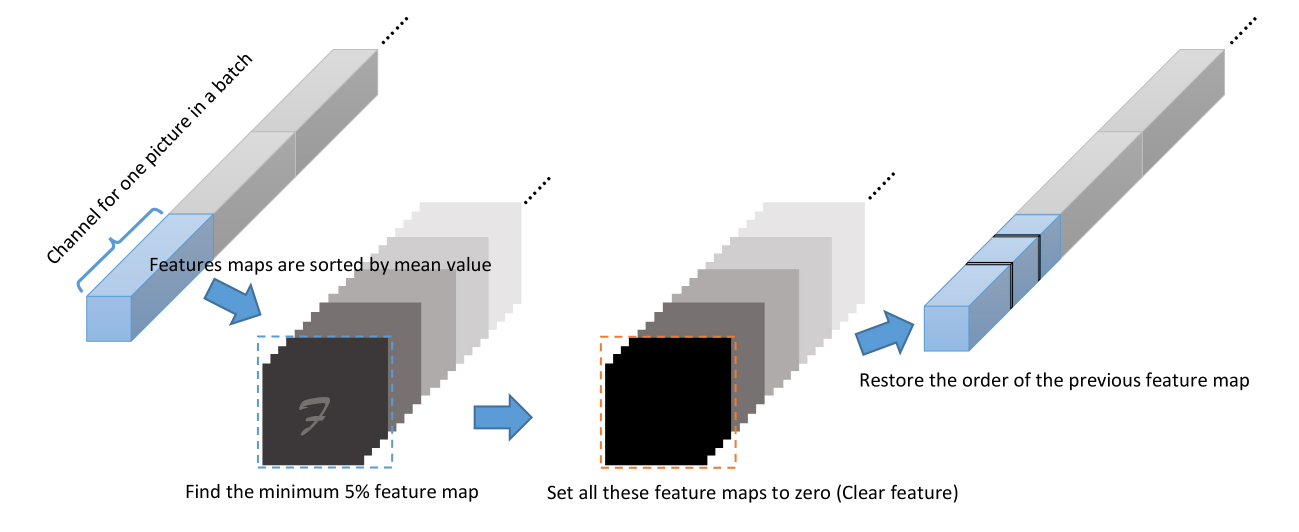}
\caption{Interference feature filtering process.}
\label{fig:iff}
\end{figure}

The quality of the feature maps produced by the convolutional network during the feature extraction process can be variable. Such variability may impact the effectiveness of deeper convolutional networks. To address this issue, we devised a module that scores each channel's feature map based on its mean value. Subsequently, we removed channel features with lower scores, where the removal ratios ranged from 0.5\% to 5\%(Fig.~\ref{fig:iff}). The proposed structure focuses on two key aspects: the output of the backbone network and the concatenation of source features and processed features (Fig. ~\ref{fig:adfp}). Interference Feature Filtering(IFF) is applied to the source feature output of the backbone network to ensure the purity of the features provided to the neck\cite{yolo4}. The belief is that the quality of source features directly impacts the quality of the processed features, which, in turn, affects the accuracy of detection.

Additionally, IFF is utilized at the concatenation of source features and processed features because the number of feature channels after concatenation is significantly larger than the number of channels that ultimately pass through CSP. The intention is to reduce low-quality features for large-scale channel compression, thereby enhancing the quality of the features output after channel compression. The percentage of filtered features decreases linearly from 5\% to 0.5\% during the training process, as the interference features decrease. It is worth noting that if too many features are filtered, it may lead to a decrease in detection accuracy.

\subsection{Adaptive Multi Positives}

\begin{figure}[H]
\centering
\includegraphics[height=7cm]{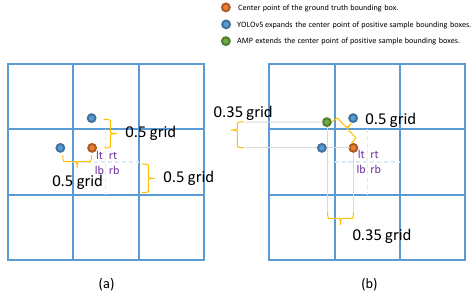}
\caption{Panel (a) illustrates the YOLOv5 method of adding positive samples, while panel (b) presents our method of adding positive samples. In panel (b), the positive sample (green dot) was added in the upper-left corner. The Euclidean distance from the new positive sample to the center of the ground truth is 0.5 grid, the same as that used by YOLOv5 to expand the positive sample.}
\label{fig:amp}
\end{figure}

Both YOLOv5\cite{yolo5} and YOLOX\cite{yolox} proposed different methods to increase the number of positive samples. YOLOv5 selects one or two positive samples from the top, bottom, left, and right grids based on the offset of the ground truth's center point, resulting in a total of two or three positive samples. YOLOX considers the grid around the center point of the ground truth as positive samples. Our proposed Adaptive Multi Positives (AMP) method is an improvement of the YOLOv5's approach (Fig.~\ref{fig:amp} (a)). The method of adding positive samples\cite{yolo5} in Fig.~\ref{fig:amp} (b) represents the first step of our proposed method. If the offset of the ground truth is greater than or equal to 0.35 grid, the positive samples in the upper left corner will not be added. With the exception of the part shown in the figure, the expansion principle for the upper right, lower right, and lower left grids remains the same, calculated based on the offset of the ground truth's center point. For instance, when the x-axis offset of the ground truth's center point is greater than 0.65 grid and the y-axis offset is less than 0.35 grid (the value of 0.35 is calculated using the Euclidean distance formula with the assumption that the X coordinate is equal to the Y coordinate, and the Euclidean distance is 0.5), the positive sample in the upper right corner is added.The calculation process is summarized as follows:

\begin{equation}
\begin{aligned}
P^0_{lt} = \{(x, y), (x-0.5, y), (x, y-0.5), (x-0.35, y-0.35) | \\
0 \leq x < 0.35 \cap 0 \leq y < 0.35 \}
\label{eq:amp1}
\end{aligned}
\end{equation}
\begin{equation}
\begin{aligned}
P^1_{lt} = \{(x, y), (x-0.5, y), (x, y-0.5) | 0.35 \leq x < 0.5 \cap 0.35 \leq y < 0.5 \}
\label{eq:amp2}
\end{aligned}
\end{equation}

\begin{equation}
\begin{aligned}
P^0_{rt} = \{(x, y), (x+0.5, y), (x, y-0.5), (x+0.35, y-0.35) | \\ 
0.65 \leq x < 1 \cap 0 \leq y < 0.35 \}
\label{eq:amp3}
\end{aligned}
\end{equation}

\begin{equation}
\begin{aligned}
P^1_{rt} = \{(x, y), (x-0.5, y), (x, y-0.5) | 0.5 \leq x < 0.65 \cap 0.35 \leq y < 0.5 \}
\label{eq:amp4}
\end{aligned}
\end{equation}

\begin{equation}
\begin{aligned}
P^0_{lb} = \{(x, y), (x-0.5, y), (x, y+0.5), (x-0.35, y+0.35) | \\
0 \leq x < 0.35 \cap 0.65 \leq y < 1 \}
\label{eq:amp5}
\end{aligned}
\end{equation}

\begin{equation}
\begin{aligned}
P^1_{lb} = \{(x, y), (x-0.5, y), (x, y+0.5) | 0.35 \leq x < 0.5 \cap 0.5 \leq y < 0.65 \}
\label{eq:amp6}
\end{aligned}
\end{equation}

\begin{equation}
\begin{aligned}
P^0_{rb} = \{(x, y), (x+0.5, y), (x, y+0.5), (x+0.35, y+0.35) | \\
0.65 \leq x < 1 \cap 0.65 \leq y < 1 \}
\label{eq:amp7}
\end{aligned}
\end{equation}

\begin{equation}
\begin{aligned}
P^1_{rb} = \{(x, y), (x-0.5, y), (x, y+0.5) | 0.5 \leq x < 0.65 \cap 0.5 \leq y < 0.65 \}
\label{eq:amp8}
\end{aligned}
\end{equation}
where p denotes the center coordinates (x or y) of the ground truth bounding box, as well as the center coordinates of the expanded positive sample bounding box. The superscript 0 indicates the application of our proposed adaptive expansion of multiple positive samples (Adaptive Multi Positives, AMP) in that condition, while superscript 1 indicates the application of YOLOv5's original method for expanding multiple positive samples in that condition. Since the method for expanding positive samples is determined based on the center coordinates (x or y) in different regions within a grid, we divide a grid into four regions: lt, rt, lb, and rb, as shown in Fig.~\ref{fig:amp}.

When the center point coordinates fall within the lt region (Fig.~\ref{fig:amp}), the expansion of positive samples is determined based on conditions for x and y. If both x and y are less than 0.35, positive samples are expanded according to Eq.\ref{eq:amp1} (Fig.~\ref{fig:amp} (b)); otherwise, they are expanded according to Eq.\ref{eq:amp2} (Fig.~\ref{fig:amp}(a)). If the center point coordinates are in the rt region, positive samples are expanded based on Eq.\ref{eq:amp3} or Eq.\ref{eq:amp4}. For coordinates in the lb region, positive samples are expanded according to Eq.\ref{eq:amp5} or Eq.\ref{eq:amp6}, while for coordinates in the rb region, positive samples are expanded using Eq.\ref{eq:amp7} or Eq.\ref{eq:amp8}.


An additional step in our approach automatically selects and augments the count of positive samples based on the dimensions of the ground truth bounding box. No positive samples are added if the ground truth bounding box size is smaller than one grid. When the ground truth bounding box size is equal to or exceeds one grid, we follow the approach employed by YOLOv5 to increase positive samples. Furthermore, if the ground truth bounding box size is equal to or exceeds two grids, we employ our proposed method to add positive samples. We avoid adding positive samples when the number of grids encompassed by the ground truth bounding box is less than one, as these additional grids lack object features. Consequently, this precaution mitigates potential false detections and their adverse impact on the network's ability to learn from the features present in the ground truth.

\subsection{Eliminate Output Sensitivity}


\begin{figure}[H]
\centering
\includegraphics[height=9cm]{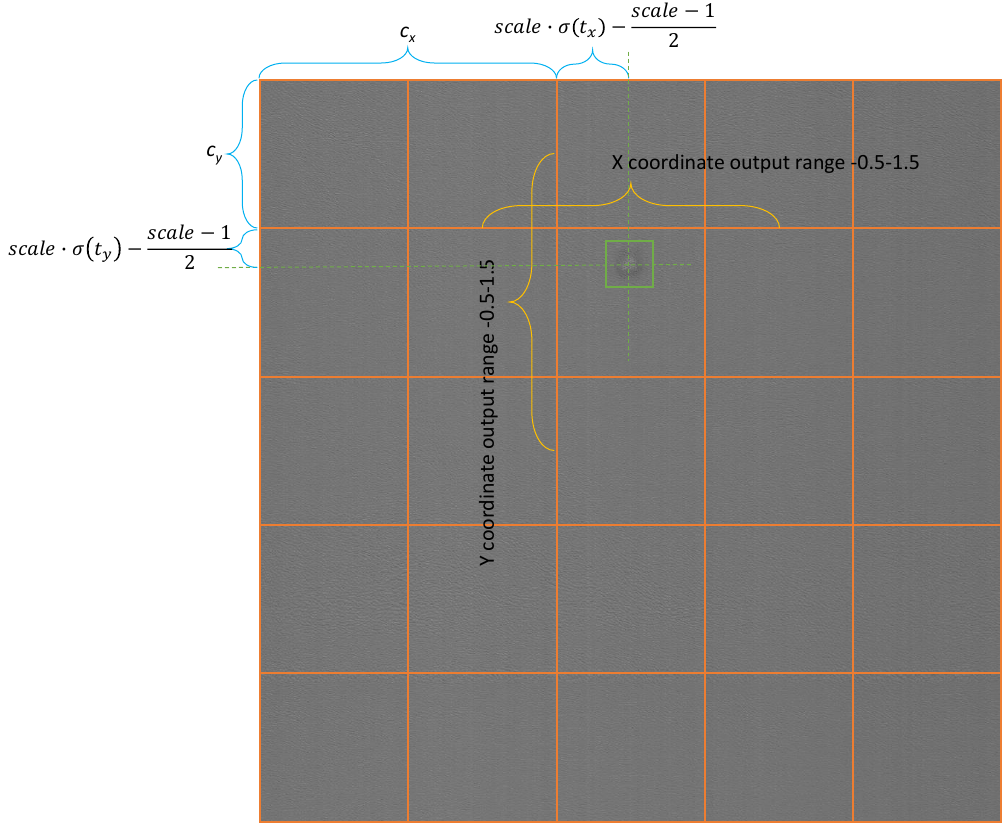}
\caption{Demonstration of the Computational Logic in Eq.\ref{eq:egs1} using a real defect image.}
\label{fig:eos}
\end{figure}

\begin{figure}[H]
\centering
\includegraphics[height=6cm]{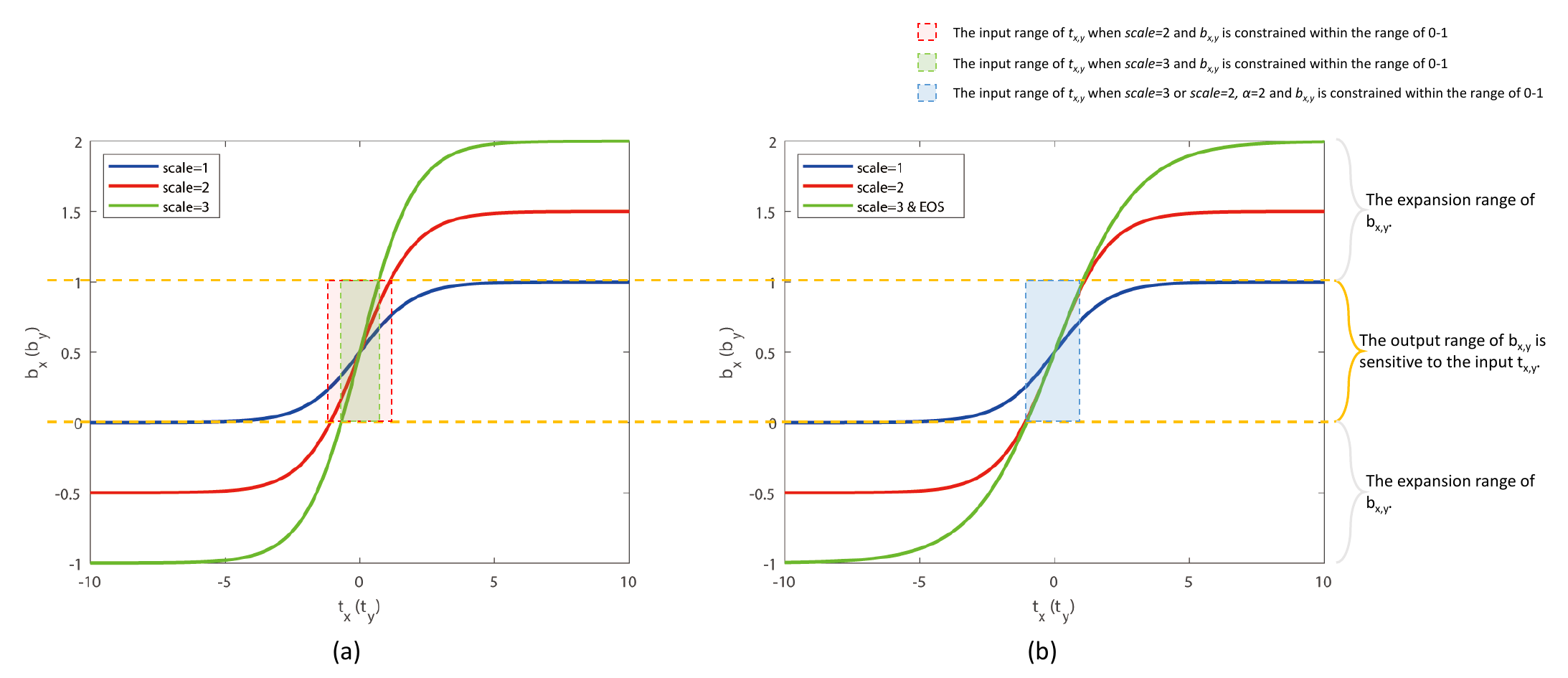}
\caption{Figure (a) shows the sigmoid function curve when scale takes different values, and figure (b) shows the sigmoid function curve when scale = 3 and EOS is added.}
\label{fig:is}
\end{figure}

Since the introduction of the method to eliminate grid sensitivity\cite{yolo4} (Eq.\ref{eq:egs1}) in YOLOv4, numerous YOLO series algorithms have adopted this approach to address the regression issues associated with bounding box center points near grid edges. However, the adoption of this method has given rise to a new problem, namely an increase in the sensitivity of the output, making it challenging for the network to converge. Therefore, we propose a method to eliminate output sensitivity(EOS). Eq.\ref{eq:egs1} introduces the concept of eliminating grid sensitivity proposed by YOLOv4. Since the coordinates of the center point of the bounding box in the network output (relative to the x or y coordinate within a grid, as shown in Fig.\ref{fig:eos}) are constrained to the range of 0-1 through the sigmoid function, this precisely corresponds to one grid. However, during the actual output of coordinates, the center point coordinates may be close to the boundary, resulting in infinite or infinitesimal inputs and causing regression deviation of the bounding box (The blue curve depicted in Fig.\ref{fig:is}(a)). To address this issue, YOLOv4 proposes the concept of eliminating grid sensitivity, summarized in Eq.\ref{eq:egs1}. In this equation, $scale$ is a scaling parameter, and in YOLOv4, it takes the value of 2. Setting $scale$ to 2 extends the output range of the bounding box center point coordinates to -0.5 to 1.5. This eliminates the need for infinite or infinitesimal inputs for outputs between 0-1.

Eq.\ref{eq:egs2} represents the derivative of Eq.\ref{eq:egs1}, aiming to articulate the significant impact of $scale$ on the slope of Eq.\ref{eq:egs1}. If we intend to further expand the range of outputs through the sigmoid function (for example, setting scale to 3), the increased slope of the output will inevitably lead to convergence issues during the network training process (Fig.\ref{fig:is}(a)).Therefore, Eq.\ref{eq:eos1} introduces an $\alpha$ parameter on the basis of Eq.\ref{eq:egs1} to separate the slope of the output equation from $scale$. This allows us to simultaneously adjust the output range through $scale$ and address the convergence issue by tuning the slope with the $\alpha$ parameter(Eq.\ref{eq:eos2}). In Eq.\ref{eq:eos1}, $\alpha$ is a constant we introduce to ensure that the slope of the output equation is solely influenced by this constant.


\begin{align}
b_{x,y} &= scale \cdot \sigma (t_{x,y}) - \frac{scale -1}{2} + c_{x,y} \label{eq:egs1} \\
b_{x,y}^{'} &= \frac{scale \cdot e^{-t_{x,y}}}{(1+e^{-t_{x,y}})^2} \label{eq:egs2} \\
b_{x,y} &= scale \cdot \sigma (t_{x,y} \cdot \frac{\alpha}{scale}) - \frac{scale -1}{2} + c_{x,y} \label{eq:eos1} \\
b_{x,y}^{'} &= \frac{\alpha \cdot e^{-t_{x,y} \frac{\alpha}{scale} }}{(1+e^{-t_{x,y} \frac{\alpha}{scale} })^2} \label{eq:eos2}
\end{align}
where $b_{x,y}$ represents the center coordinates of the bounding box output from the final layer of the network, $t_{x,y}$ is the input data from the upper-layer network structure in the head network, $scale$ is the scaling factor for eliminating grid sensitivity (with YOLOv4 taking a value of 2), $\sigma$ is the sigmoid activation function,  $c_{x,y}$ is the grid offset of the center coordinates, and $\alpha$ is a constant introduced by us to control the slope.

Fig.\ref{fig:is}(a) illustrates the curves of Eq.\ref{eq:egs1} with varying values of $scale$ (as indicated in the legend of Fig.\ref{fig:is}(a)) and $c_{x,y}$ set to 0. When the output value $b_{x,y}$ is within the range of 0 to 1 (as depicted in the orange dashed region in Fig.\ref{fig:is}), there are variations in the values of $t_{x,y}$ and changes in the slope of the curve (in the red or green dashed regions in Fig.\ref{fig:is}). It can be observed from Fig.\ref{fig:is}(a) that as $scale$ increases, the range of $t_{x,y}$ decreases, and the slope increases (as indicated by Eq.\ref{eq:egs2}, where the slope is directly related to $scale$). This directly results in larger adjustments to weight changes during the backward propagation process, making it challenging for the network to converge near the target values under the same baseline hyperparameters. Therefore, we introduce $\alpha$ in Eq.\ref{eq:egs1} (as shown in Eq.\ref{eq:eos1}), ensuring that the slope is only influenced by the fixed value of $\alpha$(as depicted in Eq.\ref{eq:eos2}), independent of $scale$. By adjusting both $\alpha$ and $scale$ parameters separately, for example, setting $\alpha$ to 2 and $scale$ to 3 (as depicted by the green curve in Fig.\ref{fig:is}(b)), we can achieve a larger range of $b_{x,y}$ values (-1 to 2), beneficial for the Adaptive Multi-Positives method. Simultaneously, we keep the input range and slope of $t_{x,y}$ constant when $b_{x,y}$ values are within the range of 0 to 1 (as indicated in the blue dashed region in Fig.\ref{fig:is}(b)), facilitating easier convergence of the network.

\section{Experiments(ALD)}
This experiment utilizes the ALD dataset to compare the detection capabilities of YOLOD and YOLOv5-r5.0\footnote{https://github.com/ultralytics/yolov5/releases/tag/v5.0} on artificial leather defects. The experimental model did not utilize any pre-trained model, and initialization parameters were randomly set. For fairness, all models were trained for only 300 epochs using mixed-precision training\cite{mpt}. Since all defects in the artificial leather dataset are single-channel gray images, conventional data augmentation methods such as mosaic\cite{yolo4}, mixup\cite{mixup}, and copy-paste\cite{copypaste} were not used in the experiment. To ensure comparability with previous models, all hyperparameters\footnote{https://github.com/ultralytics/yolov5/blob/master/data/hyps/hyp.scratch-high.yaml}, Backbone, SiLU activation, and Cosine annealing scheduler\cite{sgdr} were kept consistent with those of YOLOv5-r5.0. All experiments were conducted on a single graphics card (RTX3090). 

\begin{table}[h]
\small
\caption{Comparison of YOLOD and YOLOv5 in terms of AP on ALD. All the models are tested at 640*640 resolution, with FP16-precision and batch=1 on a RTX3090}
\label{tab:alde}
\begin{tabular}{llllll}
\hline
Models & AP$_{50}$ & AE & Parameters(M) & GFLOPs & Latency(ms) \\
\hline
YOLOv5-S & 64.4 & 20.8 & 7.3 & 17.1 & 8.7 \\
YOLOD-S & 77.9(\textcolor{green}{+13.5})& 15.6(\textcolor{red}{-5.2}) & 10.6 & 22.9 & 11.2 \\
\hline
YOLOv5-M & 78.5 & 14.4 & 21.4 & 51.4 & 11.1 \\
YOLOD-M & 90.7(\textcolor{green}{+12.2})& 7.8(\textcolor{red}{-6.6}) & 26.6 & 58.7 & 16.2 \\
\hline
YOLOv5-L & 84.6 & 10.4 & 47.1 & 115.6 & 13.7 \\
YOLOD-L & 96.3(\textcolor{green}{+11.7})& 3.2(\textcolor{red}{-7.2}) & 52.1 & 118.1 & 21.2 \\
\hline
\end{tabular}
\end{table}


Through comparative experiments (Tab.\ref{tab:alde}), YOLOD's AP$_{50}$ is significantly higher than YOLOv5, with a difference of 11.7\% - 13.5\%. This data indicates that YOLOD has a much better detection capability for artificial leather defects compared to YOLOv5. Additionally, YOLOD's AE data is 5.2\% - 7.2\% lower than YOLOv5, which demonstrates that YOLOD has a much better error detection capability for artificial leather defects than YOLOv5. In practical projects, customers also pay more attention to these two metrics. Therefore, these two metrics fully illustrate that YOLOD outperforms YOLOv5 in terms of artificial leather defect detection performance.

\section{Experiments(MS-COCO)}

The experiment employed MS-COCO 2017\cite{mscoco} dataset for verifying YOLOD, using YOLOv5-r5.0 data as the baseline. The experimental model did not utilize any pre-trained model, and initialization parameters were randomly set. For fairness, all models were trained for only 300 epochs using mixed-precision training. During training, data augmentation methods, such as mosaic, mixup, and copy-paste, were applied. To ensure comparability with previous models, all hyperparameters, Backbone, SiLU activation, Cosine annealing scheduler, and data augmentation methods were kept consistent with those of YOLOv5, except for using a smaller learning rate. Mosaic, mixup, and copy-paste were removed during training up to the 286th epoch, and all data augmentation was removed during training up to the 296th epoch. All experiments were conducted on a single graphics card (RTX3090).

\subsection{Ablation Experiment}

The experiment primarily compared the innovations' data on MS-COCO val2017. We assessed the impact of adding DFP, IFF, AMP, and EOS on the model's AP and particularly focused on the AP$_S$ (Tab.~\ref{table:table2}). As shown in Tab.\ref{table:table2}, we sequentially added the four innovative modules—IFF, AMP, EOS, and DFP—on top of the baseline model. The improvements in AP and AP$_S$ (indicated by the green numbers) demonstrate the effectiveness of these modules within the model. Specifically, adding IFF resulted in a 0.1\% increase in AP compared to the baseline. Although the addition of AMP did not yield an increase in AP compared to the baseline + IFF, it led to a 0.3\% improvement in AP$_S$. This suggests that the AMP module is effective in enhancing the model’s ability to detect fine defects. Adding EOS increased AP by 0.2\% compared to the baseline + IFF + AMP. Finally, the addition of DFP further increased AP by 0.1\% and AP$_S$ by 1.1\% compared to the baseline + IFF + AMP + EOS.



\begin{table}[h]
\small
\caption{Ablation Experiment on MS-COCO val2017. The green numbers indicate the increase in AP or AP$_S$ relative to the previous module when added to the baseline.}
\label{table:table2}
\setlength{\tabcolsep}{9mm}{
\begin{tabular}{lll}
\hline\noalign{\smallskip}
Methods & AP(\%) & $AP_S$ \\
\noalign{\smallskip}
\hline
\noalign{\smallskip}
Baseline   & 48.2 & - \\
+IFF & 48.3 (\textcolor{green}{+0.1}) & 30.9 \\
+AMP & 48.3 & 31.2 (\textcolor{green}{+0.3}) \\  
+EOS & 48.5 (\textcolor{green}{+0.2}) & 31.2  \\  
+DFP & 48.6 (\textcolor{green}{+0.1}) & 32.3 (\textcolor{green}{+1.1})\\ 
\hline
\end{tabular}}
\end{table}

\subsection{Comparative experiment}


After adding DFP, the model's AP exceeded that of YOLOv5, and there was a significant increase in the AP for small objects. For a more detailed comparison, we separately analyzed the AP of small objects; however, YOLOv5 only provides data for AP and AP$_{50}$. Consequently, we compared the data with YOLOv4-CSP and YOLOX, both having similar model sizes and the same input resolution. Our model outperformed the other models in terms of detection AP for small objects (Tab.~\ref{table:table3}).

\begin{table}[h]
\small
\caption{Comparison of YOLOD, YOLO-X and YOLOv4-CSP in terms of AP$_S$ on MS-COCO.}
\label{table:table3}
\setlength{\tabcolsep}{7mm}{
\begin{tabular}{lllll}
\hline\noalign{\smallskip}
Models & Size & $AP_S$ & $AP_M$ & $AP_L$ \\
\noalign{\smallskip}
\hline
\noalign{\smallskip}
YOLOv4-CSP & 640 & 28.2 & 51.2 & 59.8 \\
YOLO-X-L & 640 & 29.8 & 54.5 & 64.4 \\
YOLOD-L & 640 & \textbf{32.3}(\textcolor{green}{+2.5}) & 53.7 & 62.1 \\
\hline
\end{tabular}}
\end{table}

In the experiment, we used three model sizes: large (L), medium (M), and small (S), for comparison with YOLOv5. The comparison data included AP, parameters, GFLOPs, and latency (Tab.~\ref{table:table4}). Taking the L-size model as the benchmark, we scaled the depth and width of our M model by multiplying the depth by 0.67, which represents the number of residual blocks\cite{resnet} in CSP, and the width by 0.75, which represents the number of channels. For the S model, the depth was multiplied by 0.3, and the width was multiplied by 0.5. Our model achieved 0.4\%–2.6\% higher AP than YOLOv5, with a minimal difference in parameters and GFLOPs.

\begin{table}[h]
\small
\caption{Comparison of YOLOD and YOLOv5 in terms of AP on MS-COCO. All the models are tested at 640*640 resolution, with FP16-precision and batch=1 on a RTX3090}
\label{table:table4}

\setlength{\tabcolsep}{4mm}{
\begin{tabular}{lllll}
\hline\noalign{\smallskip}
Models & AP(\%) & Parameters(M) & GFLOPs & Latency(ms) \\
\noalign{\smallskip}
\hline
\noalign{\smallskip}
YOLOv5-S & 36.7 & 7.3 & 17.1 & 8.7 \\
YOLOD-S & \textbf{39.3}(\textcolor{green}{+2.6}) & 10.6 & 22.9 & 11.2 \\
\hline
YOLOv5-M & 44.5 & 21.4 & 51.4 & 11.1 \\
YOLOD-M & \textbf{45.6}(\textcolor{green}{+1.1}) & 26.6 & 58.7 & 16.2 \\
\hline
YOLOv5-L & 48.2 & 47.1 & 115.6 & 13.7 \\
YOLOD-L & \textbf{48.6}(\textcolor{green}{+0.4}) & 52.1 & 118.1 & 21.2 \\
\hline
\end{tabular}}
\end{table}

\subsection{Comparison with the state-of-the-art}

\begin{table}[h]
\small
\caption{Comparison of the speed and accuracy of different object detectors on MS-COCO 2017 test-dev. We select all the models
trained on 300 epochs for fair comparison.}
\label{table:table5}
\begin{tabular}{llcccccccc}
\hline
Method & Size & FPS & AP(\%) & $AP_{50}$ & $AP_{75}$ & $AP_S$ & $AP_M$ & $AP_L$ \\
\hline
\hline
RetinaNet  & 640 & 37 &  37 & - & - & - & - & - \\
RetinaNet  & 640 & 29.4 &  37.9 & - & - & - & - & - \\
RetinaNet  & 1024 & 19.6 &  40.1 & - & - & - & - & - \\
RetinaNet  & 1024 & 15.4 &  41.1 & - & - & - & - & - \\
\hline
YOLOv3 + ASFF* & 320 & 60 &  38.1 & 57.4 & 42.1 & 16.1 & 41.6 & 53.6 \\
YOLOv3 + ASFF* & 416 & 54 &  40.6 & 60.6 & 45.1 & 20.3 & 44.2 & 54.1 \\
YOLOv3 + ASFF* & 608× & 45.5 &  42.4 & 63.0 & 47.4 & 25.5 & 45.7 & 52.3 \\
YOLOv3 + ASFF* & 800× & 29.4 &  43.9 & 64.1 & 49.2 & 27.0 & 46.6 & 53.4 \\
\hline
EfficientDet-D0 & 512 & 62.5 & 33.8 & 52.2 & 35.8 & 12.0 & 38.3 & 51.2 \\
EfficientDet-D1 & 640 & 50.0 & 39.6 & 58.6 & 42.3 & 17.9 & 44.3 & 56.0 \\
EfficientDet-D2 & 768 & 41.7 & 43.0 & 62.3 & 46.2 & 22.5 & 47.0 & 58.4 \\
EfficientDet-D3 & 896 & 23.8 & 45.8 & 65.0 & 49.3 & 26.6 & 49.4 & 59.8 \\
\hline
PP-YOLOv2 & 640 & 68.9 & 49.5 & 68.2 & 54.4 & 30.7 & 52.9 & 61.2 \\
PP-YOLOv2 & 640 & 50.3 & 50.3 & 69.0 & 55.3 & 31.6 & 53.9 & 62.4 \\
\hline
YOLOv4 & 608 & 62.0 & 43.5 & 65.7 & 47.3 & 26.7 & 46.7 & 53.3 \\
YOLOv4-CSP & 640 & 73.0 & 47.5 & 66.2 & 51.7 & 28.2 & 51.2 & 59.8 \\
\hline
YOLOX-M & 640 & 81.3 & 46.4 & 65.4 & 50.6 & 26.3 & 51.0 & 59.9 \\
YOLOX-L & 640 & 69.0 & 50.0 & 68.5 & 54.5 & 29.8 & 54.5 & 64.4 \\
\hline
YOLOv3-ultralytics & 640 & 95.2 & 44.3 & 64.6 & - & - & - & - \\
YOLOv5-S & 640 & 114.9 & 36.7 & 55.4 & - & - & - & - \\
YOLOv5-M & 640 & 90.1 & 44.5 & 63.1 & - & - & - & - \\
YOLOv5-L & 640 & 73.0 & 48.2 & 66.9 & - & - & - & - \\
\hline
YOLOD-S & 640 & 89.2 & 39.3 & 56.6 & 42.6 & 22.1 & 44.2 & 51.8 \\
YOLOD-M & 640 & 61.7 & 45.6 & 63.5 & 49.5 & 29.1 & 50.4 & 59.4 \\
YOLOD-L & 640 & 47.2 & 48.6 & 66.7 & 52.7 & \textbf{32.3} & 53.7 & 62.1 \\
\hline
\end{tabular}
\end{table}


Finally, the experiment was compared with other state-of-the-art (SOTA) data (Tab.~\ref{table:table5}). As access to a Tesla V100 graphics card was unavailable, the model's calculation speed depended on hardware and software, beyond our control. Therefore, FPS calculations were conducted using the available RTX3090 graphics cards, and FPS was calculated without considering the time taken for post-processing and NMS.

Upon comparing the model with other SOTA approaches, it was observed that the calculation speed was close to that of YOLOv5, while the model's accuracy had been improved to a certain extent. Notably, the detection accuracy of the model for small objects exceeded that of the other models.

Due to the simultaneous comparison in Tab.~\ref{table:table5} of the AP values for different-scale models of the baseline, the SOTA models, and the next-generation algorithm YOLOX (which was the latest algorithm at that time), our proposed model YOLOD (L, M, S-scale models) demonstrated comprehensive superiority over both the baseline and the SOTA models (excluding YOLOX) in terms of AP. Furthermore, the AP$_S$ of YOLOD-L not only surpassed YOLOX-L but also exceeded the AP$_S$ of YOLOX-X (where X denotes a larger-scale model than L). In addition, the focus of this paper is on the improvement of detection for fine defects in artificial leather. Since the object size of these fine defects aligns with the object size in AP$_S$, the detection metric of AP$_S$ is particularly crucial for assessing the performance of our proposed YOLOD in detecting fine defects. Therefore, in Tab.~\ref{table:table5}, the AP$_S$ data for YOLOD-L outperforms that of other state-of-the-art models used for comparison.

\section{Conclusion}



In this study, four effective innovation modules were proposed for the structure of the YOLOv5 model, combined with the defect characteristics of artificial leather, forming a new algorithm called YOLOD. YOLOD achieves a higher AP(0.4\% - 2.6\%) than YOLOv5 on the general MS-COCO dataset, with minimal impact on the inference speed. Additionally, YOLOD exhibits better AP (+11.7\% to +13.5\%) and lower AE (-5.2\% to -7.2\%) than YOLOv5 on the artificial leather defect dataset. Addressing the persistent challenge of small object detection in YOLO series algorithms, the method proposed in this paper effectively resolves the issue, resulting in a notable enhancement in AP$_S$ (+2.5\% and +4.1\% compared to YOLOX-L and YOLOv4-CSP, respectively). Moreover, it performs well in detecting small defects in artificial leather, providing a useful reference for other researchers in applying YOLO series models more effectively.In future work, further optimization of the YOLOD model will focus on enhancing its adaptability to a broader range of defect types and different materials. Additionally, exploring more innovative modules to improve its performance in detecting more complex and subtle defects will help expand its application across various industrial domains.
﻿
﻿


\section{Acknowledge}
This work was supported by the National Natural Science Foundation of China [Nos. 62331008, 62027827, 62221005 and 62276040], 
Natural Science Foundation of Chongqing (Nos. 2023NSCQ-LZX0045 and CSTB2022NSCQ-MSX0436).





\bibliography{yolod}

\end{document}